\documentclass[pmlr,twocolumn,10pt, letterpaper]{jmlr} 

\usepackage{arydshln}
\usepackage{booktabs}
\usepackage{multirow}
\usepackage{hyperref}
\usepackage{adjustbox}
\usepackage{svg}
\usepackage{pifont}
\newcommand{\xmark}{\ding{55}}%
\usepackage{siunitx}

\usepackage[switch]{lineno}

\theorembodyfont{\upshape}
\theoremheaderfont{\scshape}
\theorempostheader{:}
\theoremsep{\newline}

\jmlrvolume{248}
\jmlryear{2024}
\jmlrsubmitted{LEAVE UNSET}
\jmlrpublished{LEAVE UNSET}
\jmlrworkshop{Conference on Health, Inference, and Learning (CHIL) 2024} 

\title[Interpretation of Intracardiac Electrograms Through Textual Representations]{Interpretation of Intracardiac Electrograms \\Through Textual Representations}

\author{%
\Name{William Jongwon Han} \Email{wjhan@andrew.cmu.edu}\\
\addr Carnegie Mellon University, USA
\AND
\Name{Diana Gomez} \Email{dggomez@andrew.cmu.edu}\\
\addr Carnegie Mellon University, USA
\AND
\Name{Avi Alok} \Email{aavi@andrew.cmu.edu}\\
\addr Carnegie Mellon University, USA
\AND
\Name{Chaojing Duan} \Email{chaojind@andrew.cmu.edu}\\
\addr Allegheny Health Network, USA
\AND
\Name{Michael A. Rosenberg} \Email{michael.a.rosenberg@cuanschutz.edu}\\
\addr University of Colorado, USA
\AND
\Name{Douglas Weber} \Email{dweber2@andrew.cmu.edu}\\
\addr Carnegie Mellon University, USA
\AND
\Name{Emerson Liu} \Email{emersonliu@msn.com}\\
\addr Allegheny Health Network, USA
\AND
\Name{Ding Zhao} \Email{dingzhao@andrew.cmu.edu}\\
\addr Carnegie Mellon University, USA
}

\begin{document}

\maketitle

\begin{abstract}
Understanding the irregular electrical activity of atrial fibrillation (AFib) has been a key challenge in electrocardiography.
For serious cases of AFib, catheter ablations are performed to collect intracardiac electrograms (EGMs).
EGMs offer intricately detailed and localized electrical activity of the heart and are an ideal modality for interpretable cardiac studies. 
Recent advancements in artificial intelligence (AI) has allowed some works to utilize deep learning frameworks to interpret EGMs during AFib.
Additionally, language models (LMs) have shown exceptional performance in being able to generalize to unseen domains, especially in healthcare.
In this study, we are the first to leverage pretrained LMs for finetuning of EGM interpolation and AFib classification via masked language modeling.
We formulate the EGM as a textual sequence and present competitive performances on AFib classification compared against other representations.
Lastly, we provide a comprehensive interpretability study to provide a multi-perspective intuition of the model's behavior, which could greatly benefit the clinical use.

\end{abstract}

\paragraph*{Data and Code Availability}
We recorded intracardiac electrograms in the left atrium of two patients, one healthy and the other afflicted with atrial fibrillation (AFib), via an Octoray catheter from Biosense Webster Inc.
The recording was done at Allegheny General Hospital in Pittsburgh, Pennsylvania.
Our recorded data will not be released. 
We also experiment with the Intracardiac Atrial Fibrillation Database \citep{goldberger_2000_physiobank} accessible through this link: \href{https://physionet.org/content/iafdb/1.0.0/}{https://physionet.org/content/iafdb/1.0.0/}.
They used a decapolar catheter with 7mm spacing between bipoles to record the right atria from 8 patients in atrial fibrillation or flutter.
More details about the data is provided in Section~\ref{exp}.
We have released the code  at the following link: \href{https://github.com/willxxy/Text-EGM}{https://github.com/willxxy/Text-EGM}.
\paragraph*{Institutional Review Board (IRB)}
An official IRB document has been received from Allegheny Health Network deeming our work with the collected data as exempt status under the Code of Regulations: 45 CFR 6.104 (d) Exempt 4. The Intracardiac Atrial Fibrillation Database is publicly available and de-identified, thus not needing an IRB approval.

\section{Introduction}
\label{sec:intro}
Atrial fibrillation (AFib) is one of the most common types of arrhythmia globally, affecting more than 60 million people over the last thirty years \citep{elliott_2023_epidemiology}.
AFib is characterized by the chaotic, irregular, and often rapid, beating pattern of the heart's upper chambers, known as the atria. 
This causes the atria to become out of sync with the heart's lower chambers, called the ventricles. 
Due to these debilitating properties, people with AFib are at risk of stroke, heart failure, and a number of other cardiac diseases. 

\begin{figure*}[htp]
\centering
\includegraphics[width=1\linewidth]{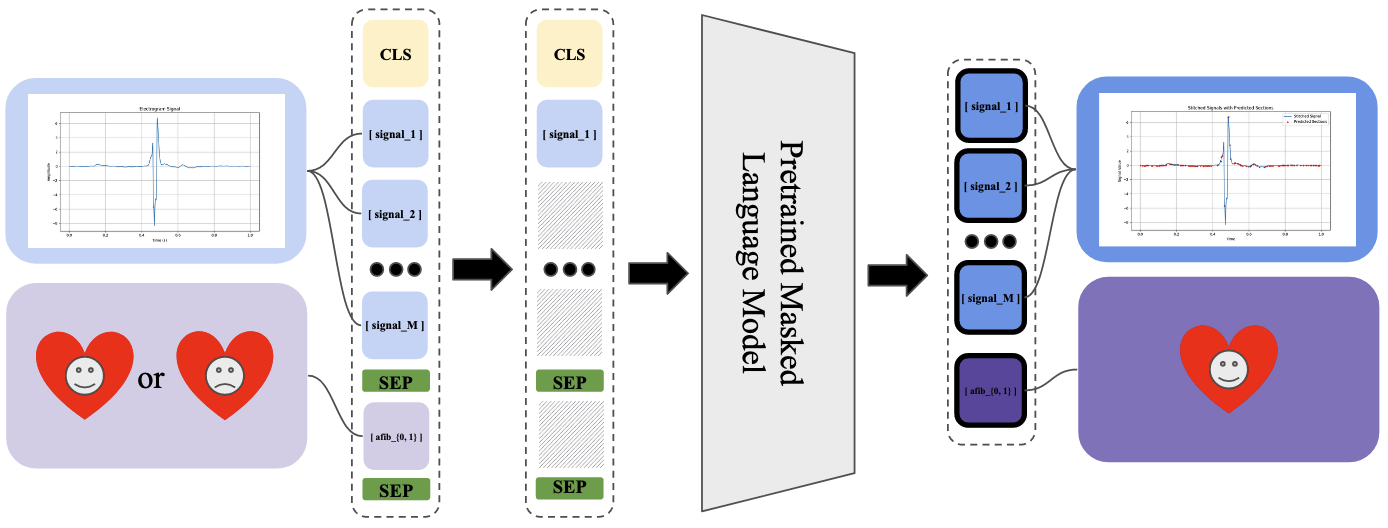}
\caption{The overall pipeline of our model. We formulate EGM signal interpolation and AFib classification as a masked language modeling task. \includegraphics[width=0.25cm]{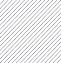} denotes the mask.}
\label{Fig:model}
\end{figure*}

Catheter ablations are performed for patients with more serious cases of AFib, during which intracardiac electrograms (EGMs) representative of electrical activity along the endocardial surface of the atria are collected by means of navigational catheters embedded with multiple electrodes. 
Because EGMs have distinct and spatially specific signatures for different types of cardiac arrhythmia, they are the ideal modality for highly interpretable cardiac studies, in particular for AFib. 

Due to their locality and richness in information, there are works utilizing deep learning methods with EGMs for AFib classification \citep{alhusseini_2020_machine}, patient-specific therapy \citep{john_2022_artificial}, and surface electrocardiogram (ECG) reconstruction \citep{zhang2021rtrcg}.
In addition, the representation of EGMs are carefully considered \citep{alhusseini_2020_machine, tang_2022_machine} and are most commonly represented as visual or time series modalities \citep{e25020332, zhang2021rtrcg, alhusseini_2020_machine}. 
Despite the impressive performances of these EGM algorithms, the model's interpretability is a critical component that requires further exploration before deployment in the clinical setting. 
Although some works attempt to interpret the model's decisions, to our best knowledge, only a uniperspective interpretability metric is provided, such as visualizing the Grad-CAM heatmap \citep{alhusseini_2020_machine}, which can often be misleading \citep{ribeiro2016why, donosoguzmán2023comprehensive}. 

In this study, we introduce a tokenization schema, inspired by \citet{chen2022pix2seq}, that represents an EGM as a textual sequence.
The tokenization process discretizes continuous EGM signal amplitudes and maps each to a unique token ID, similarly to textual data.
Representing EGM signals as a textual sequences allows us to leverage powerful Language Models (LMs) pretrained on billions of texts.
Specifically, we use a Masked Language Model (MLM) that is optimized for predicting randomly masked positions in a given sequence.
As seen in Figure~\ref{Fig:model}, the AFib label is also tokenized input.
By masking 75\% of the signal tokens and the AFib label token, the model is able to interpolate the masked EGM values and simultaneously classify the EGM as AFib or not.
Additionally, we provide a multi-perspective interpretability procedure to expose the intuition behind the model's decisions at the token level. 

In summary, our main contributions are the following:

\begin{itemize}
    \item To our best knowledge, this is the first work to represent EGMs as a textual sequence. We introduce an effective tokenization schema that maintains the low level information of the original signal. 
    \vspace{-1mm}
    \item We utilize a MLM pretrained on \textbf{textual data} to finetune for interpreting EGM signals by interpolation and classification for AFib.
    \vspace{-1mm}
    \item We also perform a comprehensive interpretability procedure via attention maps, integrated gradients \citep{sundararajan2017axiomatic}, and counterfactual analysis to provide clarity of the model's decisions for clinicians. 
    \vspace{-3mm}
\end{itemize}

\section{Related Work}
\label{sec:rel}
\subsection{Language Models for Healthcare}
Language Models (LMs) have been an extremely popular medium for many different downstream tasks in healthcare. 
\citet{qiu-etal-2023-transfer} utilizes LMs to translate surface ECG signals into descriptive clinical notes.
\citet{choi2023ecgbert} first cluster surface ECG signals into 70 groups and uses them to create a wave vocabulary for pretraining a MLM. 
They then employ the pretrained MLM model for various downstream tasks, such as AFib classification, heartbeat classification, and user identification \citep{choi2023ecgbert}.
There has also been large efforts in pretraining LMs on solely clinical text, such as \citet{alsentzer2019publicly} and \citet{li2022clinicallongformer}.
\citet{mehandru2023large} observed that as the role of LMs become larger in the medical world, they can act as agents by providing clinical decision support and stakeholder interactions. 
Therefore, they propose new evaluation frameworks, termed Artificial-intelligence Structured Clinical Examinations (AI-SCI), to assess LMs in real-world clinical tasks \citep{mehandru2023large}.

\subsection{Machine Learning for EGM}
Machine learning in electrogram (EGM) analysis, though limited, has seen notable advancements. 
\citet{alhusseini_2020_machine} used a 64-electrode catheter to create 8x8 spatial activation heatmaps from atrial EGM signals, feeding them into a CNN for atrial fibrillation (AFib) classification. 
\citet{zhang2021rtrcg} developed RT-RCG, a neural network tailored for ECG data reconstruction from EGMs, incorporating a Differentiable Acceleration Search for efficient hardware accelerator optimization. 
\citet{DUQUE2017182} presented a genetic algorithm and K-NN based classifier to categorize EGM signals, enhancing AFib ablation therapy guidance. 
\citet{tang_2022_machine} employed a multimodal approach, integrating surface ECG, EGM signals, and clinical data in a CNN, improving AFib classification accuracy. 
Our study extends these works by applying LMs, specifically MLMs, pretrained on general text to interpret complex EGM signals in AFib cases.

\vspace{-3mm}
\subsection{Interpretability for Machine Learning in Healthcare}
Interpretability in healthcare machine learning is crucial but challenging for clinical use \citep{healthex}. 
\citet{jin2021explainable} outline four attribution methods for machine learning interpretability: backpropagation, feature perturbation, attention, and model distillation. 
Notably, \citet{alhusseini_2020_machine} and \citet{vaid2022heartbeit} use Grad-CAM heatmaps and attention saliency maps, respectively, to interpret CNN and Vision Transformer models. 
\citet{shi_2022_learning} advocates for interpretability in healthcare models, with \citet{sanchez_2022_causal} and \citet{prosperi_2020_causal} emphasizing counterfactual analysis and addressing biases in machine learning. 
Our study merges several interpretability methods (e.g., attention, integrated gradients, counterfactual analysis) for a more holistic understanding of model decisions.
\vspace{-5mm}
\section{Methods}
\label{sec:method}
Our method consists of three main steps: (1) translating the EGM signals from time series to text through tokenization, (2) utilizing MLMs to predict the masked portions of the input sequence, and (3) interpreting the decisions of the models through attention weights, integrated gradients, and counterfactual analysis.
\vspace{-3mm}
\subsection{Tokenization}

Our tokenization approach, inspired by \citet{chen2022pix2seq}, discretizes the continuous amplitudes of EGM signals to create a textual representation of the time series information. In this section, we intricately explain the process of tokenization for a single sequence. Let $S = \{s_1, s_2, \ldots, s_M\} \in \mathbb{R}^{M}$ be a continuous EGM signal, where $M=1000$ for a one second signal.

The normalization of $S$ scales the amplitude values to [0, 1]. For the sequence $S$ and each $s_i$, this is given by:
\[
S_{\text{norm}} = \left\{ \frac{s_i - s_{\text{min}}}{s_{\text{max}} - s_{\text{min}}} : s_i \in S \right\}
\]
where $s_{\text{min}} = \min(S)$, $s_{\text{max}} = \max(S)$, and $S_{\text{norm}}$ is the normalized sequence.

Quantization converts $S_{\text{norm}}$ into discrete levels $V$, where $V = 250$.
The quantized sequence $S_{\text{quant}}$ is:
\[
S_{\text{quant}} = \left\{ \left\lfloor S_{\text{norm}} \times V \right\rfloor : s_i \in S_{\text{norm}} \right\}
\]
where $\left\lfloor \cdot \right\rfloor$ denotes the floor function.
The decision of the value of $V$ were purposely chosen through experimentation. 
We observed that $250$ was the minimum amount of discrete levels that represents the fine-grained details of an EGM signal as well as obtain the best performance for the signal interpolation task.

We then map $S_{\text{quant}}$ to a unique token ID and expand the tokenizer's embedding table by adding the new tokens. For signals, we define:
\[
T_{\text{S}} = \left\{ \text{``signal\_"} + \text{str}(q) : q \in S_{\text{quant}} \right\}
\]
where $\text{str}(\cdot)$ converts numbers to strings. 
We also create the token IDs for the AFib label by introducing the following tokens:
\[
T_{\text{A}} = \left\{ \text{``afib\_"} +\text{str}(a) : a \in {0, 1}\right\}
\]
where 0 and 1 denotes a normal and AFib signal respectively.
\vspace{-3mm}
\subsection{Masked Language Model}
The Masked Language Models (MLMs) we use are BigBird \citep{zaheer2021big}, LongFormer \citep{beltagy2020longformer}, Clinical BigBird \citep{li2022clinicallongformer}, and Clinical LongFormer \citep{li2022clinicallongformer}, which are optimized to handle sequences of lengths up to 4,096, for finetuning for EGM signal interpolation and AFib classification.
Clinical BigBird and Clinical Longformer uses the same architecture as their predecessors, however, they are pretrained on 2 million clinical notes from the MIMIC-III dataset \citep{Johnson2016MIMICIIIAF} instead of a general web-crawled text dataset.
More details on the models are available in the appendix.
\vspace{-3mm}
\subsection{Masking Strategy}
To prepare our input for the MLM, we adopt a high masking strategy.
We randomly mask out $75\%$ of $T_S$. 
We choose $75\%$, as suggested in \citet{he2021masked}, to ensure the model learns a significant portion of the EGM waveform and to diminish the possibility of interpolating from the surrounding unmasked areas.
We always mask out $T_A$ to ensure the model is not being fed the classification label.
After applying the masks, our input is formulated as follows:
\[
T_I = [\text{CLS}] \oplus T_S \oplus [\text{SEP}] \oplus T_A \oplus [\text{SEP}]
\]
where $\oplus$ is the concatenation operation, [\text{CLS}] is the classification token indicating the start of the sequence, and [\text{SEP}] is the separator token denoting both the end of a sequence as well as distinguishing boundaries for different types of input \citep{devlin2019bert}.
\vspace{-3mm}
\subsection{Learning Objectives}
\paragraph{Masked Language Model Loss} Let \( M \) be the set of indices of the masked tokens in \( T_I \). For each masked token index \( i \in M \), the original token at this position is \( t_i \). The MLM objective is to minimize the following loss function:

\[
L_{MLM}(\theta) = -\sum_{i \in M} \log P(t_i | T_I; \theta),
\]

\noindent where \( \theta \) represents the parameters of the MLM model, \( P(t_i | T_I; \theta) \) is the probability predicted by the model for the token \( t_i \) given the masked input sequence \( T_I \), and the sum is over all masked tokens in the sequence.

\paragraph{Atrial Fibrillation Classification Loss} 
We add another loss $L_{AFib}$ based on the prediction of \( T_{A} \) within the sequence \( T_I\). The cross-entropy loss for this classification task is computed as follows:
\[
L_{AFib}(\theta) = -\sum_{c=1}^{C} y_{c} \log(p_{c}),
\]
where \( C \) is the number of classes for AFib classification, \( y_{c} \) is a binary indicator (0 or 1) if class label \( c \) is the correct classification for \( T_{A} \), and \( p_{c} \) is the predicted probability for the token \( T_{A} \) being of class \( c \).

\paragraph{Final Learning Objective}
The total loss for the model is given by:
\[
L(\theta) = \alpha_1 L_{MLM}(\theta) + \alpha_2 L_{AFib}(\theta),
\]
where \(\alpha_i \in [0,1], i \in \{1,2\}\).

\vspace{-3mm}
\subsection{Model Interpretability}
All of our model interpretability procedures have been conducted using the BigBird \citep{zaheer2021big} model with the final learning objective.

\paragraph{Attention}
Visualizing the attention map is an extremely common technique for interpretability in Transformers \citep{vaswani2023attention}.
There has been arguments in the interpretability community that debate whether attention maps are reliable or not \citep{jain2019attention, wiegreffe2019attention}.
The general consensus seems to be that although attention can be informative, it can also be misleading if it is the only metric of interpretability \citep{wen2023transformers}.
Therefore, we visualize the attention maps of our models, while providing other perspectives as well.
In our study, we visualize the averaged attention map over all heads and layers. 

\paragraph{Integrated Gradients}
Integrated Gradients ($IG$), introduced by \citep{sundararajan2017axiomatic}, is a method for attributing a model's prediction to its input features, providing insight into the model's behavior. 
Given an input \( x \) and a baseline \( x' \), $IG$ computes the gradient of the model's output with respect to the input along the path $\alpha$ from the baseline to the input. 
The attribution of each feature \( i \) is given by:
\[
IG_i(x) = (x_i - x'_i) \times \int_{\alpha=0}^{1} \frac{\partial F(x' + \alpha \times (x - x'))}{\partial x_i} d\alpha 
\]
where \( F \) is the model function. $IG$ satisfies two key axioms: Sensitivity, ensuring non-zero attribution for features that change the output, and Implementation Invariance, guaranteeing consistent attributions for functionally equivalent models.
In our setting, the input $x$ is our masked, tokenized sequence, $T_I$.
The baseline $x'$ is defined to be a vector of padding tokens [PAD] of the same dimension as $T_I$. 

\paragraph{Counterfactual Analysis} Counterfactual analysis observes the change of the predicted outcome by directly manipulating the input \citep{Feder_2021, Pearl09}.
We individually conduct three different counterfactual analysis methods: Token Substitution, Token Addition, and Label Flipping.
We view the effects of the introduced counterfactuals in two settings. 
The first setting is where we finetune the MLM without the counterfactuals and inference with modified and unmodified inputs. 
The second setting is where we finetune the MLM with the counterfactuals alongside unmodified inputs, and inference on both modified and unmodified inputs. 
In both settings, we randomly choose 25\% of the batch to be modified with the counterfactuals.

\paragraph{Token Substitution}
Given our input signal $T_S$, we apply token substitution by replacing the original sequence with a smoothed out version via a moving average filter.
The intuition behind this is aligned from the clinicians' perspective such that the sharp oscillations in a given signal is where most of the information is stored for the model or clinician to determine whether it is a normal heartbeat or AFib.
By smoothing out the oscillations via a moving average filter, we want to observe the model's robustness against this counterfactual as well as its attention and gradients. 

\paragraph{Token Addition}
Given our input signal $T_S$, we apply token addition by introducing the following $V$ new augmentation signal tokens and adding them to tokenizer's embedding table:
\[
T_{\text{AUG}} = \left\{ \text{``augsig\_"} + \text{str}(q) : q \in S_{\text{quant}} \right\}
\]
Please note that $T_{\text{AUG}}$ is still derived from the same $S_{\text{quant}}$ as $T_S$.
We then randomly choose 25\% (a 250 length segment) of $T_{\text{AUG}}$ and append it to $T_S$ therefore getting the input sequence 
\[
T_I' = [\text{CLS}] \oplus T_S \oplus T_{\text{AUG}} \oplus [\text{SEP}] \oplus T_A \oplus [\text{SEP}]
\]
These augmented tokens will not be masked.
The ground truth label to these augmented tokens will be the padding token [PAD]. 
Essentially, the augmented tokens will simply serve as noise to $T_S$ to observe whether the model leverages the noise or still attends to the meaningful, non-augmented portion of the signal. 
\paragraph{Label Flipping}
Given our masked input AFib label $T_A$, we apply label flipping by replacing the ground truth label with its opposite. 
This is used to understand the model's behavior under adversarial examples.

\begin{table*}[hbtp]
\floatconts
  {tab:main_results}
  {\caption{AFib classification results with different baselines and representations.}}
  {
  \begin{center}
  \begin{adjustbox}{width=1\linewidth}
  \begin{tabular}{lccccccc}
  \toprule
  \bfseries Method
  & \bfseries Representation
  & \bfseries Sensitivity \%
  & \bfseries Specificity \%
  & \bfseries PPV \%
  & \bfseries NPV \%
  & \bfseries Accuracy \%
  \\
  \midrule
  CatBoost \citep{tang_2022_machine} & Time Series &88.5 & 62.7 & - & - & 70.1\\
  CNN \citep{alhusseini_2020_machine} & Image &97.0 &	93.0 & 93.1 & 97.0 & 95.0\\
  K-Means \citep{alhusseini_2020_machine} & Image&77.0 & 82.3 & 83.5	& 75.5 & 79.4\\
  KNN \citep{alhusseini_2020_machine} & Image & 75.3 & 84.0 & 87.1 & 70.3 & 78.9\\
  LDA \citep{alhusseini_2020_machine} & Image & 85.0 & 74.6 & 76.4 & 83.7 & 79.7\\
  SVM  \citep{alhusseini_2020_machine} & Image & 82.9 & 76.7 & 77.4 & 82.3 & 79.7 \\
  \midrule
  ViT \citep{dosovitskiy2021image} & Image &\textbf{100.0} & 99.1 & 99.1 & \textbf{100.0} & \textbf{99.7} \\
  BEiT \citep{bao2022beit} & Image & 99.9 & 99.3 & 98.8 & 99.9 & 99.5 \\
  BigBird \citep{zaheer2021big} & Time Series & 62.2 & 68.7 & 65.3 & 63.5 & 63.6\\
  LongFormer \citep{beltagy2020longformer} & Time Series & 60.6 & 62.0 & 64.2 & 61.2 & 64.6\\
  BigBird \citep{zaheer2021big} & Text & 97.8 & 96.1 & 93.5 & 98.7 & 96.7\\
  LongFormer \citep{beltagy2020longformer} & Text & 69.0 & 98.4 & 96.1 & 84.7 & 87.7\\
  Clinical BigBird \citep{li2022clinicallongformer} & Text & 12.8 & 97.3 & 72.8 & 66.1 & 66.6\\
  Clinical LongFormer \citep{li2022clinicallongformer} & Text & 85.2 & 97.9 & 95.8 & 92.1 & 93.3\\
  \midrule
  \textbf{Ours BigBird \citep{zaheer2021big}} & Text & 96.7 & \textbf{99.8} & 99.6 & 98.2 & 99.2 \\
  \textbf{Ours LongFormer \citep{beltagy2020longformer}} & Text & \textbf{100.0} & 99.4 & 99.0 & 99.8 & 99.5\\
  \textbf{Ours Clinical BigBird \citep{li2022clinicallongformer}} & Text & 99.0 & 99.0 & 98.1 & 99.4 & 99.0\\
  \textbf{Ours Clinical LongFormer \citep{li2022clinicallongformer}} & Text & 99.9 & 99.6 & \textbf{99.9 }& 99.3 & \textbf{99.7}\\
  \bottomrule
  \end{tabular}
  \end{adjustbox}
  \end{center}
  }
\end{table*}

\begin{figure}[htp]
\centering
\includegraphics[width=1\linewidth]{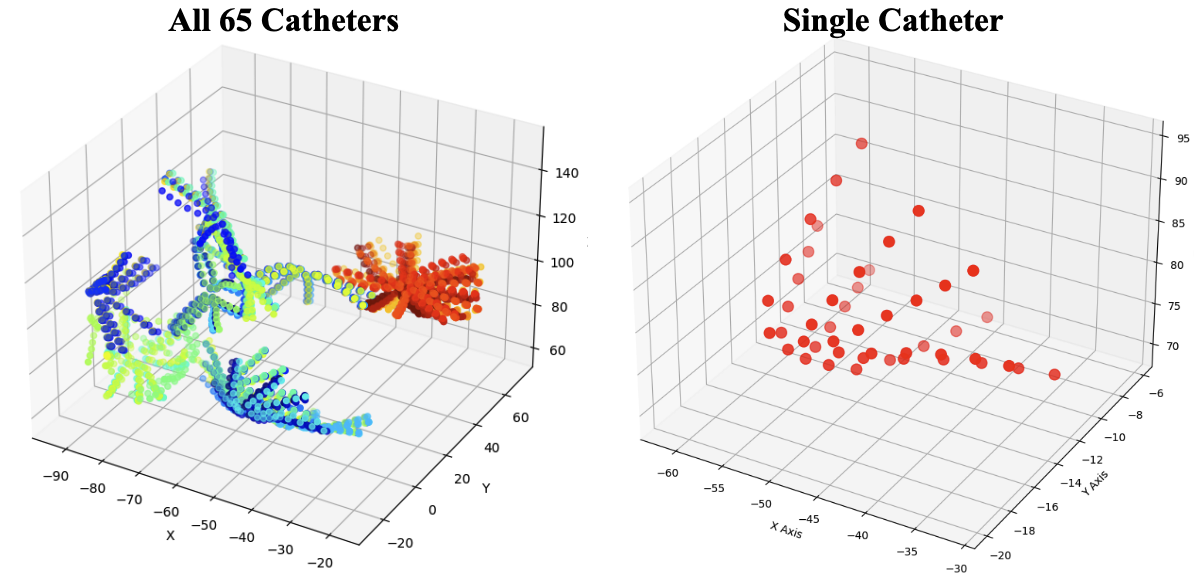}
\caption{Visualization of all 65 (left) Octoray catheters and a single Octoray Catheter (right) placements inside the left atrium.}
\label{Fig:cath}
\end{figure}
\vspace{-5mm}
\section{Experiments}
\label{exp}
\subsection{Dataset and Preprocessing}
Our dataset consists of 20 different placements of catheter ablation for a patient with normal heartbeat rhythm and 45 different placements of catheter ablation for a patient with AFib.
The catheter ablation was performed with the Octoray catheter from Biosense Webster Inc.
The catheter consists of 8 splines with 6 electrodes on each spline, summing up to 48 electrodes.
The catheter ablation performed on the patient with AFib was for 30 seconds and for the patient with a normal heartbeat, the ablation was for 29 seconds.
Both studies were sampled at a rate of 1000 Hz.
It is important to note that our dataset is from two patients.
However, this limitation is overcome by the diversity of catheter placement locations as seen in Fig~\ref{Fig:cath}.
We treat each one-second signal from each electrode on all 65 catheter placements as a separate instance.
In summary, we have a total number of $(29 * 48 * 20) + (30 * 48 * 45) = 92,640$ unique one-second samples (27,840 from normal heartbeat and 64,800 from AFib) that we use for finetuning.

Additionally, we finetune and inference on the Intracardiac Atrial Fibrillation Database \citep{goldberger_2000_physiobank} to show the generalizability of our method to external datasets.
This dataset consists of endocardial recordings from the right atria of 8 patients in atrial fibrillation or flutter \citep{goldberger_2000_physiobank}.
They utilize a decapolar catheter with a 7 millimeter spacing between bipoles at a sampling rate of 1000 Hz.
Four separate catheter placements of the heart were recorded, and for each region, 5 bipolar signals were recorded.

Let $\mathbf{X} \in \mathbb{R}^{I \times J \times K}$ represent the normal and AFib data tensor, with dimensions $I, J$ and $K$ denoting the time length, number of electrodes, and number of catheter placements, respectively. Z-score normalization is applied across the recording time and electrodes for each placement $k$. The normalized tensor $\mathbf{Z}$ is computed as:
\[
Z_{i, j, k} = \frac{X_{i, j, k} - \mu_{k}}{\sigma_{k}}
\]
where $\mu_{k}$ and $\sigma_{k}$ are the mean and standard deviation of the elements in $\mathbf{X}$ across the first two dimensions for each $k$.
We then segment $Z_{i,j,k}$ into a sequence of non-overlapping segments of length $M$, arriving at $S = \{s_1, s_2, \ldots, s_M\}$.
Unless specified otherwise, we report the results where $M = 1000$.

\begin{table*}[hbtp]
\floatconts
  {tab:intra}
  {\caption{Results on inferencing on the external dataset \citep{goldberger_2000_physiobank} where we finetuned with our dataset ($\clubsuit$) or external dataset ($\diamondsuit$) for AFib Classification and interpolation.}}
  {\small \begin{tabular}{lccccccccccc}
  \toprule
  \multirow{2}{*}{\bfseries Model}
  & \multirow{2}{*}{\bfseries Finetuned}
  & \multicolumn{2}{c}{\bfseries Interpolation}
  & \multicolumn{1}{c}{\bfseries AFib Classification} 
  \\
  &&   \bfseries MSE $\downarrow$ 
  & \bfseries MAE $\downarrow$ 
  & \bfseries Accuracy \%
  \\
  \midrule 
  \textbf{Ours BigBird \cite{zaheer2021big}} & $\clubsuit$ & 0.77 & 0.29 & 80.3\\
  \textbf{Ours LongFormer \cite{beltagy2020longformer}} & $\clubsuit$ & \textbf{0.75} & 0.29 & 74.08 \\ 
  \textbf{Ours Clinical BigBird \cite{li2022clinicallongformer}} & $\clubsuit$ & 0.80 & 0.23 & \textbf{81.8}\\
  \textbf{Ours Clinical LongFormer \cite{li2022clinicallongformer}} & $\clubsuit$ &  \textbf{0.75} & \textbf{0.21} & 73.83 \\ 
  \midrule
  \textbf{Ours BigBird \cite{zaheer2021big}} & $\diamondsuit$ & \textbf{0.86 }& 0.30 & 99.8\\
  \textbf{Ours LongFormer \cite{beltagy2020longformer}} & $\diamondsuit$ & 1.02 & 0.25 & 99.8\\ 
  \textbf{Ours Clinical BigBird \cite{li2022clinicallongformer}} & $\diamondsuit$ & \textbf{0.86} & \textbf{0.24} & \textbf{99.9}\\
  \textbf{Ours Clinical LongFormer \cite{li2022clinicallongformer}} & $\diamondsuit$ &  1.02 & 0.26 & 99.7 \\ 
  \bottomrule
  \end{tabular}}
\end{table*}

\subsection{Experimental Setting}
We finetuned the masked language model utilizing the AdamW optimizer \citep{kingma2017adam} with a learning weight of $1e-4$, weight decay rate of $1e-2$ \citep{adamw}.
We conducted all experiments with a batch size of 8 and 1 for finetuning and inference, respectively. 
For our final learning objective $L$, $\alpha_1 = \alpha_2 = 1$. 
Our experiments were conducted on 2 NVIDIA A6000 and 2 NVIDIA A5000 GPUs. 

During inference, we evaluate our models on two tasks, namely EGM signal interpolation and AFib classification.
For EGM signal interpolation, we use the Mean Squared Error and Mean Absolute Error. 
For AFib classification, we use the sensitivity, specificity, positive predicted values (PPV), negative predicted values (NPV), and accuracy metrics for evaluation. Additionally, all reported results are utilizing our own dataset unless specified otherwise.

\section{Results}
\label{sec:results}
\subsection{AFib Classification}
We compare our results with different representations and baselines in \tableref{tab:main_results}.
We divide our table into three sections. 
The top section contains prior works that do AFib classification with EGMs. 
It is important to note that the datasets used in this top section are \textbf{different from ours} (i.e., different patients, catheter, sampling rate), therefore they are not directly comparable. 
However, we decide to include them in this table to report the current state of AFib classification using EGMs.

The middle section contains the results from different base models and representations with our dataset.
For ViT \citep{dosovitskiy2021image} and BEiT \citep{bao2022beit}, we transform the EGMs into image representations via Markov Transition Field (MTF), Gramian Angular Field (GAF) \citep{Wang2014EncodingTS}, and Recurrence Plot (RP) \citep{recplot}, which have proven to be excellent image representations of time series signals \citep{pmlr-v225-qiu23a}.
We load in the pretrained weights of ViT and BEiT, `google/vit-base-patch16-224-in21k' and `microsoft/beit-base-patch16-224-pt22k' respectively, and apply the 75\% masking rate on the images during finetuning.
For the time series representation results using LongFormer \citep{beltagy2020longformer} and BigBird \citep{zaheer2021big}, we simply utilize the normalized signal amplitudes as input embeddings by projecting them through a linear layer before passing them into their respective models.
For the text representation results, we report the results for only utilizing the pretrained base model with masked language model loss. 

In the bottom section of the table, we report our results for utilizing the textual representation and full learning objective. 
We can observe that our method of representing the complex EGM signal as a simple textual sequence achieves competitive results with the image modality.

Lastly, we report the results of AFib classification on the Intracardiac Atrial Fibrillation Database \citep{goldberger_2000_physiobank} in Table~\ref{tab:intra}. 
Table~\ref{tab:intra} shows the AFib classification results under two settings: 1) we finetune on our own collected dataset and inference on the Intracardiac Atrial Fibrillation Database \citep{goldberger_2000_physiobank} and 2) we finetune and inference on the external dataset.
Observably, simply inferencing on the external dataset after finetuning on our dataset yields competitive results, with Clinical BigBird \citep{li2022clinicallongformer} achieving 81.8\% accuracy.
When we finetune our method on the external dataset, we can see that our method achieves near perfect results, with Clinical BigBird \citep{li2022clinicallongformer} achieving 99.9\% accuracy.
These results emphasize the generalizability of our method to external datasets under the two settings.

\begin{table}[!hbtp]
\floatconts
  {tab:inter}
  {\caption{Results on interpolation.}}
  {\resizebox{\columnwidth}{!}{%
  \centering \small
  \begin{tabular}{lcccccc}
  \toprule
  \multirow{2}{*}{\bfseries Method}
  & \multicolumn{2}{c}{\bfseries Interpolation}
  \\
  &  \bfseries MSE $\downarrow$ 
  & \bfseries MAE $\downarrow$ 
  \\
  \midrule
  BigBird \citep{zaheer2021big}  & 0.44 & 0.16 \\
  LongFormer \citep{beltagy2020longformer}  & 0.37&\textbf{0.13} \\
  Clinical BigBird \citep{li2022clinicallongformer} &0.50 &0.17 \\
  Clinical LongFormer \citep{li2022clinicallongformer} & \textbf{0.36}& \textbf{0.13}\\
  \midrule
  \textbf{Ours BigBird \citep{zaheer2021big}} &0.80 &0.37 \\
  \textbf{Ours LongFormer \citep{beltagy2020longformer}} & 0.40 & 0.14 \\
  \textbf{Ours Clinical BigBird \citep{li2022clinicallongformer}} &0.44 & 0.15\\
  \textbf{Ours Clinical LongFormer \citep{li2022clinicallongformer}} &0.40 &0.14 \\
  \bottomrule
  \end{tabular}%
  }}
\end{table}
\vspace{-5mm}

\subsection{EGM Interpolation}
We report the results of the EGM interpolation task on our own dataset and the external dataset \citep{goldberger_2000_physiobank} in \tableref{tab:inter} and \tableref{tab:intra}, respectively.
We visualize the comparison of the reconstructed (blue) and ground truth (yellow) signal in the appendix.
From \tableref{tab:inter}, we can see the models with only the MLM learning objective opposed to our final learning objective achieves slightly lower MSE and MAE scores. 
We can see a clear trade-off between the interpolation results and the AFib classification results in adding the Cross Entropy (CE) loss on top of the MLM loss.
However, this trade-off is acceptable due to the increase in performance for AFib classification.

\vspace{-3mm}
\subsection{Ablation Study}
\paragraph{Pretrained vs Non-Pretrained}
We observe the effects of utilizing pretrained vs non-pretrained versions of the MLM in \tableref{tab:pt}. 
For clarification, pretrained means we load in the checkpoint that was pretrained on the large text corpus and conduct finetuning. 
Conversely, non-pretrained means we do not load the checkpoint that was pretrained on the text corpus and conduct finetuning. 
Although the BigBird model \citep{zaheer2021big} and LongFormer model \citep{beltagy2020longformer} were pretrained on a text corpora that was not directly related to clinical subjects, we see that the pretrained setting greatly outperforms the non-pretrained setting.
From this, we can observe the characteristic of pretrained models being able to leverage the language knowledge to well adapt to unseen domains.

\begin{table*}[hbtp]
\floatconts
  {tab:pt}
  {\caption{Ablation study on pretrained (\checkmark) vs non-pretrained (\xmark) MLM.}}
  {\small \begin{tabular}{lccccccccccc}
  \toprule
  \multirow{2}{*}{\bfseries Model}
  & \multirow{2}{*}{\bfseries Pretrained}
  & \multicolumn{2}{c}{\bfseries Interpolation}
  & \multicolumn{1}{c}{\bfseries AFib Classification} 
  \\
  &&   \bfseries MSE $\downarrow$ 
  & \bfseries MAE $\downarrow$ 
  & \bfseries Accuracy \%
  \\
  \midrule 
  BigBird \cite{zaheer2021big} & \xmark & 5.42 & 1.42 & 63.6\\
  BigBird \cite{zaheer2021big} & \checkmark & \textbf{0.44} & \textbf{0.16} & \textbf{96.7}\\
  LongFormer \cite{beltagy2020longformer} & \xmark &6.32 & 2.21 & 60.9\\ 
  LongFormer \cite{beltagy2020longformer} & \checkmark & 0.37 &0.13 & 87.7  \\ 
  \bottomrule
  \end{tabular}}
\end{table*}
\vspace{-5mm}

\subsection{Interpretability}
\begin{figure*}[htp]\small
\centering
\includegraphics[width=0.90\linewidth]{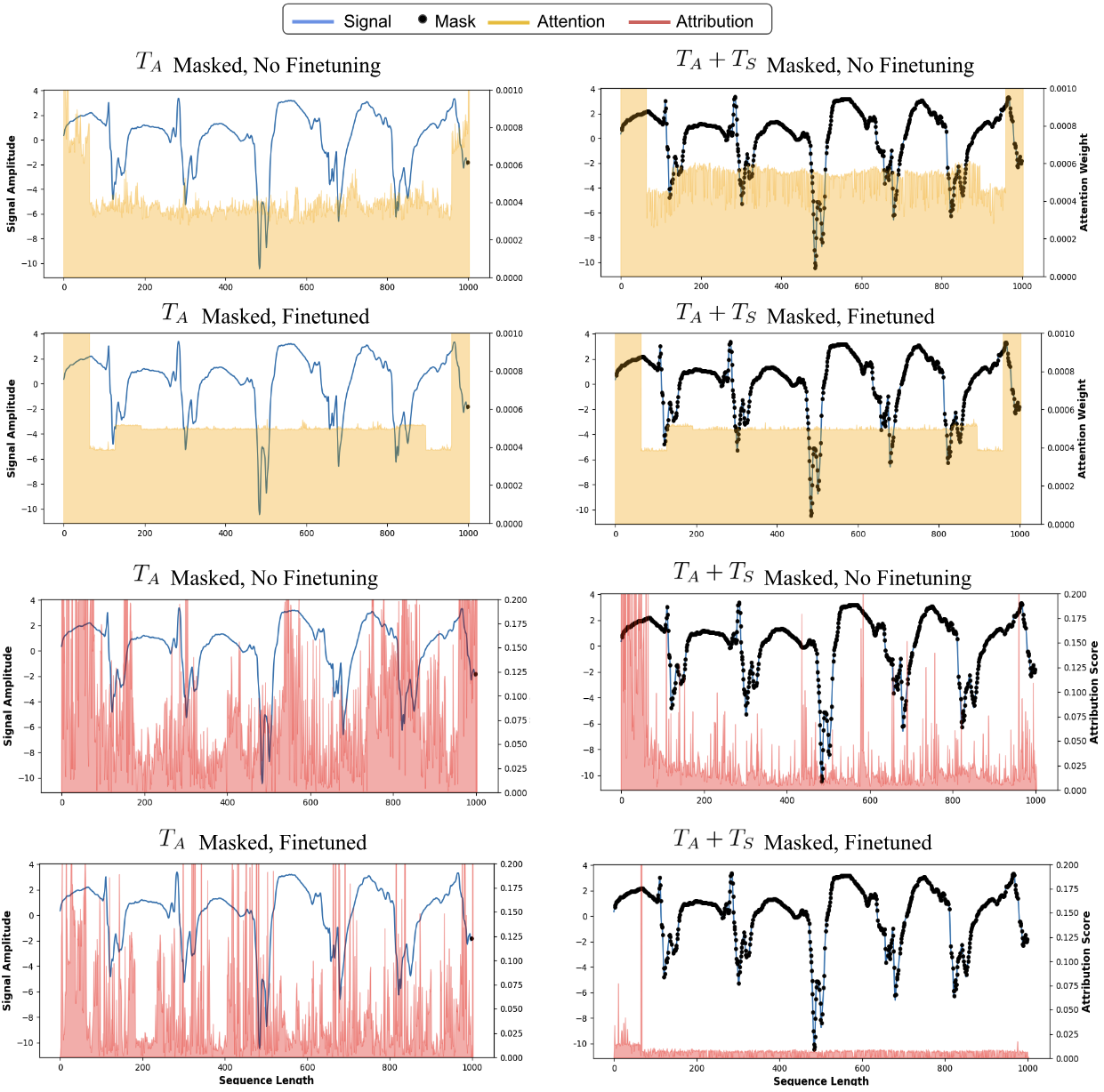}
\caption{The averaged attention weights (yellow) and attribution scores (red) of an AFib EGM signal during inference. `$T_A$ Masked' refers to running inference when only the $T_A$ token is masked. `$T_A$ + $T_S$ Masked' denotes our regular paradigm of masking out 75\% of the $T_S$ tokens and always masking out the $T_A$ token.}
\label{Fig:att_int_map}
\end{figure*}

In our study, we analyzed AFib EGM signal attention maps and attribution scores during inference (Figure~\ref{Fig:att_int_map}), examining both before and after finetuning, and under different masking conditions ($T_A$ Masked, $T_A + T_S$ Masked).

Attention maps pre-finetuning showed slightly more oscillation compared to post-finetuning, though the overall distribution remained largely unchanged in both settings and masking conditions. Notably, the model focused more on the start and end of sequences, a behavior attributed to its pretraining on natural language data, where salient information is often found at these positions. This tendency seems to have generalized to EGM data.

Regarding attribution scores, a marked difference pre- and post-finetuning was observed in score stability. Pre-finetuning, scores were random across the signal when masking only the $T_A$ token and somewhat concentrated at the start but still randomly distributed with our original masking strategy. Post-finetuning, the distribution changed significantly. 
With only the $T_A$ token masked, the scores peaked around signal oscillations, aligning with clinical significance. 
For full masking, scores were evenly distributed across the sequence, with notable attribution at the start, reflecting the MLM's use of the entire sequence for contextual understanding, including predicting the $T_A$ token. 
We visualize more random samples of attribution scores overlayed on both normal and AFib EGMs in the appendix.

\begin{figure*}[!t]\small
\centering
\includegraphics[width=0.90\linewidth]{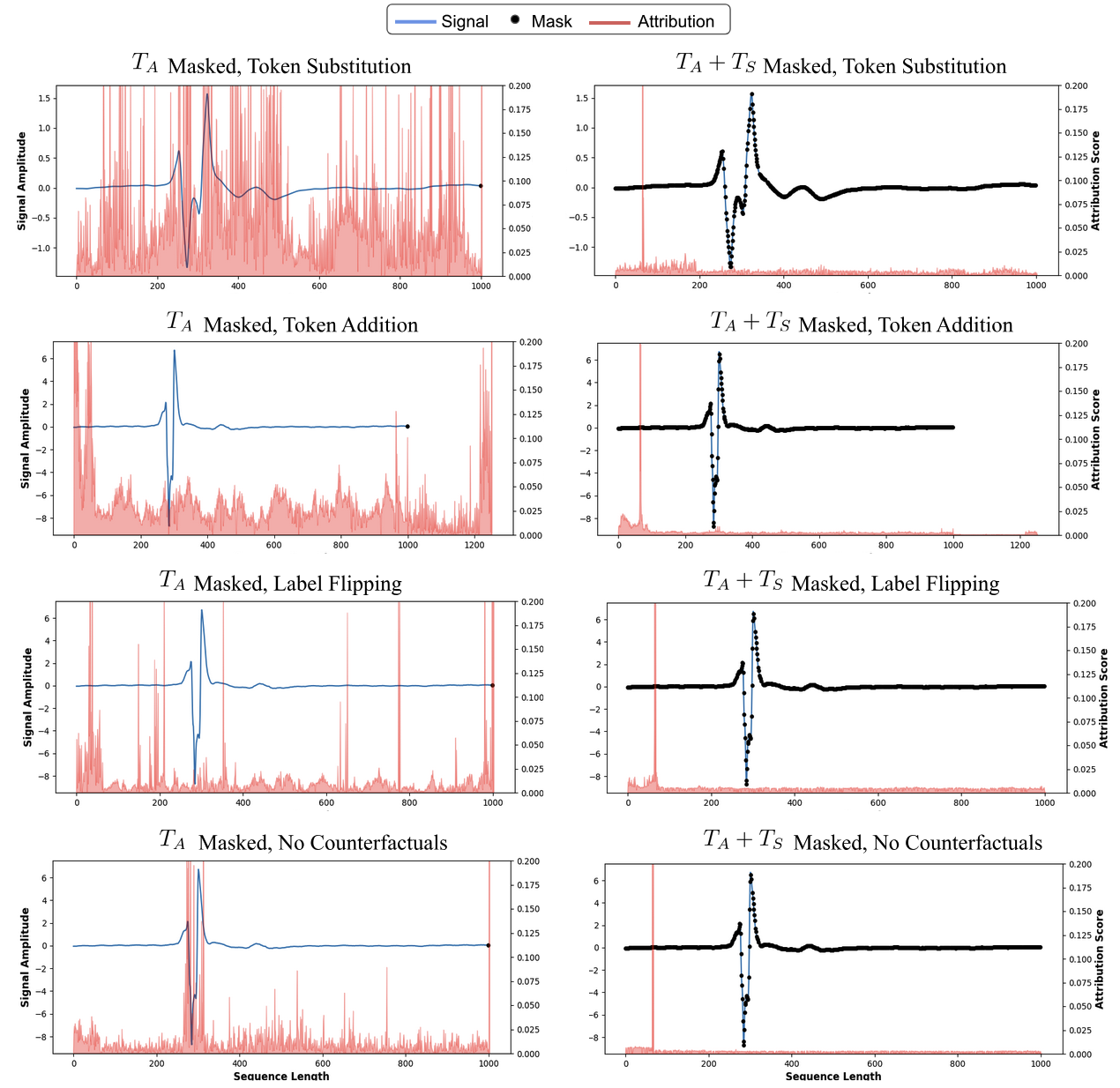}
\caption{The averaged attribution scores for checkpoints that have been finetuned on their respective counterfactuals. We also include the visualization of the average attribution score for the checkpoint that has been finetuned without counterfactuals for comparison.}
\label{Fig:cf}
\end{figure*}

\begin{table*}[hbtp]
\floatconts
  {tab:aug}
  {\caption{Results for finetuning with (\checkmark) and without (\xmark) Token Substitution, Token Addition, and Label Flipping, where we highlight the best results for each counterfactual in \textbf{\textcolor{orange}{orange}}, \textbf{\textcolor{blue}{blue}}, and \textbf{\textcolor{red}{red}}, respectively.}}
  {\small \footnotesize \begin{tabular}{lccccccccccccccc}
  \toprule
  \multirow{2}{*}{\bfseries Representations}
  & \multirow{2}{*}{\bfseries Counterfactuals}
  &\multirow{2}{*}{\bfseries Finetuned}
  & \multicolumn{2}{c}{\bfseries Interpolation}
  & \multicolumn{1}{c}{\bfseries AFib Classification} 
  \\
  &&&   \bfseries MSE $\downarrow$ 
  & \bfseries MAE $\downarrow$ 
  & \bfseries Accuracy \%
  \\
  \midrule 
  \multirow{5}{*}{Image} &\multirow{2}{*}{Token Substitution}& \checkmark & 0.57 & 0.46 &\textbf{\textcolor{orange}{99.4}}\\
  & & \xmark & 0.72& 0.55 & \textbf{\textcolor{orange}{99.0}}\\
  \cdashline{2-6}
  &\multirow{2}{*}{Token Addition}&  \checkmark & 1.12 & 0.64  &\textbf{\textcolor{blue}{99.1}} \\
  &&  \xmark &  1.92 & 1.24  & 73.8 \\
  \cdashline{2-6}
  &\multirow{2}{*}{Label Flipping}& \checkmark & 2.31 & 1.12 &74.0\\ 
  && \xmark& 2.51 & 2.08 & 74.5\\
  \midrule 
  \multirow{5}{*}{Time Series} &\multirow{2}{*}{Token Substitution}& \checkmark & 11.63& 6.40 & 37.1 \\
  & & \xmark & 11.96 & 6.34 & 36.4\\
  \cdashline{2-6}
  &\multirow{2}{*}{Token Addition}&  \checkmark & 8.81 &  5.10 & 54.2 \\
  &&  \xmark & 9.92 & 5.50  & 50.7\\
  \cdashline{2-6}
  &\multirow{2}{*}{Label Flipping}& \checkmark & 7.22 & 4.51 & 56.9\\ 
  && \xmark& 10.41 & 9.56 & 56.1\\
\midrule 
  \multirow{5}{*}{Text} &\multirow{2}{*}{Token Substitution}& \checkmark & \textbf{\textcolor{orange}{0.29}} & \textbf{\textcolor{orange}{0.26}} & 96.3\\
  & & \xmark &\textbf{\textcolor{orange}{0.25}} & \textbf{\textcolor{orange}{0.25}} & 74.7\\
  \cdashline{2-6}
  &\multirow{2}{*}{Token Addition}&  \checkmark & \textbf{\textcolor{blue}{0.70}} & \textbf{\textcolor{blue}{0.34}} & 99.0\\
  &&  \xmark &  \textbf{\textcolor{blue}{0.61}} & \textbf{\textcolor{blue}{0.19}}   & \textbf{\textcolor{blue}{77.4}} \\
  \cdashline{2-6}
  &\multirow{2}{*}{Label Flipping}& \checkmark & \textbf{\textcolor{red}{0.47}}& \textbf{\textcolor{red}{0.17}} & \textbf{\textcolor{red}{93.5}}\\ 
  && \xmark& \textbf{\textcolor{red}{0.81}} & \textbf{\textcolor{red}{0.38}} & \textbf{\textcolor{red}{91.3}}\\
  \bottomrule
  \end{tabular}}
\end{table*}
\vspace{-5mm}
\subsection{Counterfactual Analysis}
We first analyzes model performance with three counterfactuals (Token Substitution, Token Addition, Label Flipping) across two finetuning settings (i.e., with/without counterfactuals) and three representations (i.e., text, image, time series), using ViT and BigBird architectures for image and time series respectively. Textual representations exhibit higher robustness to counterfactuals in interpolation tasks compared to others.

Attribution scores from checkpoints finetuned on respective counterfactuals were compared against non-counterfactual methods (Figure~\ref{Fig:cf}). For Token Substitution, full masking results in typical attribution score distributions, while partial masking ($T_A$ token) shows random distributions, indicating a forced learning of a generalized signal distribution. Token Addition demonstrates a focused attribution on oscillating signal parts when $T_A$ is masked, reflecting model adaptation to identify key sequence elements amidst noisy tokens, enhancing interpolation and AFib classification (see \tableref{tab:aug}).

For Label Flipping, the attribution scores under full masking align with our standard method, indicating robustness to adversarial examples. However, partial masking shows arbitrary peaks across the sequence, diverging from the focused attribution of our non-counterfactual method. This suggests inherent model resilience against incorrect labels without specific adversarial training objectives.

\vspace{-5mm}
\section{Discussion and Conclusion}
In this study, we demonstrated the effectiveness of using pretrained MLMs with textual representations for EGM signal interpretation. 
Our approach yielded impressive results in AFib classification and interpolation on our own internal dataset as well as the external dataset \citep{goldberger_2000_physiobank}, achieving 99.7\% accuracy, 0.40 MSE, and 0.14 MAE, and 99.9\% accuracy, 0.86 MSE, and 0.24 MAE, respectively. 
We compared the use of a general text corpus pretrained MLM versus a non-pretrained MLM for EGM interpretation.
The pretrained model significantly outperformed the non-pretrained model in both AFib classification and interpolation, showing a 38.8\% increase in accuracy and decreases of 5.92 in MSE and 2.07 in MAE.
Our analysis included visualizing the attention and attribution scores. 
While attention weights were not insightful regarding the model's decision-making process, the attribution scores under full masking indicated substantial reliance on pretrained knowledge for interpreting EGM signals. 
Interestingly, partial masking revealed the model's focus on oscillating signal parts for AFib classification, aligning with clinical interpretations.
Further, we demonstrated our method's robustness by finetuning it with predefined counterfactuals, which showed superior performance over other representations. 
Analyzing the attribution scores of these finetuned models revealed distinct attribution distributions for each counterfactual. 
The distinct behaviors observed in the attribution scores of these finetuned models not only confirm the adaptability of our approach but also open avenues for future research. 
Specifically, these findings encourage further exploration into optimizing the model's focus on critical signal segments, maintaining accuracy even when confronted with varying counterfactual scenarios.
Additionally, this work not only contributes to the field of EGM signal interpretation via deep learning but also sets a foundation for future studies aimed at expanding the capabilities of LM-based approaches in EGM analysis.
\vspace{-2mm}
\paragraph{Limitations}
While our study highlights the potential of utilizing general pretrained MLMs for interpreting complicated EGM signals during AFib, there are notable limitations worth mentioning.
The most obvious limitation of this study is the data. 
Due to the invasive and complex nature of collecting intracardiac EGM data, it is difficult to find and conduct comparable baselines.
However, we try to mitigate this limitation in our work by conducting several baseline models with different representations and external datasets, namely the Intracardiac Atrial Fibrillation Database \citep{goldberger_2000_physiobank}. 

\label{sec:disc}
\vspace{-5mm}
\acks{\vspace{-3mm}This work is done in collaboration with the Mario Lemieux Center for Heart Rhythm Care at Allegheny General Hospital.}

\bibliography{chil-sample}

\appendix
\vspace{-3mm}
\section{Additional Visualizations}
\label{apd:first}

\begin{figure*}[!t]
\centering
\includegraphics[width=0.99\linewidth]{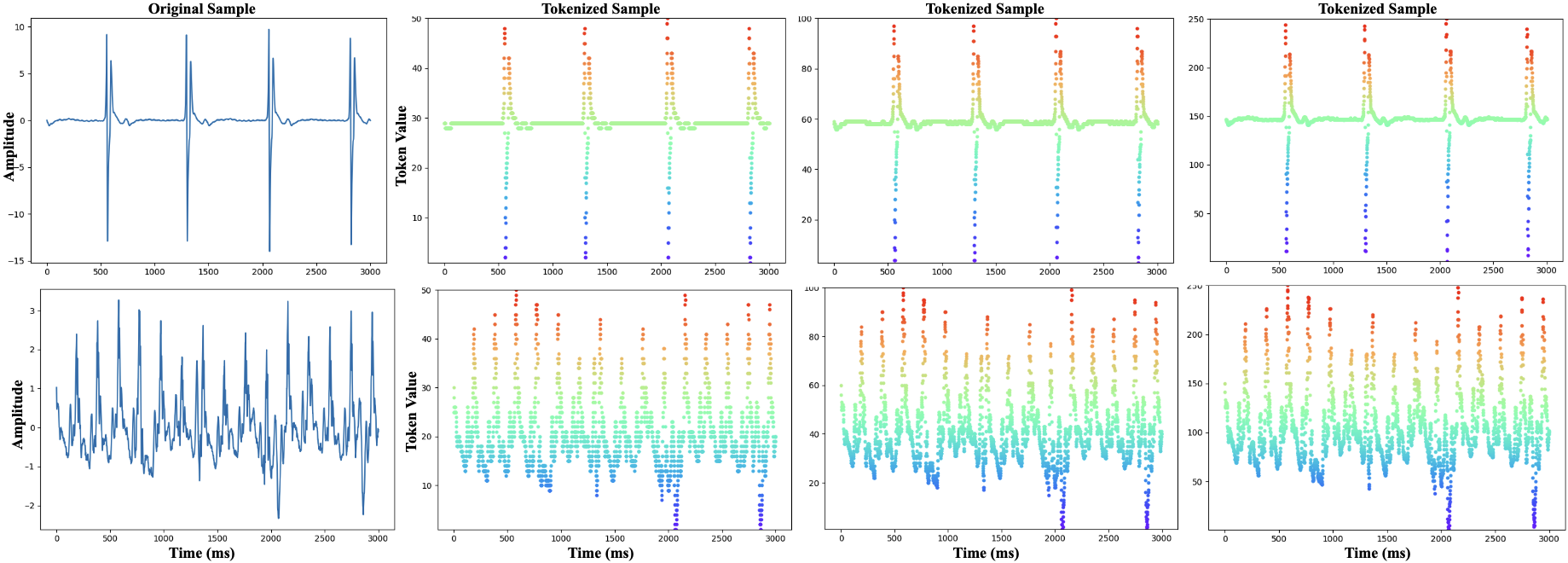}
\caption{The tokenized representation of an EGM signal for a patient with a normal heartbeat (Top) and AFib (Bottom) when $V = \{50, 100, 250\}$ levels from left to right and starting with the the original time series EGM representation.}
\label{Fig:token}
\end{figure*}

\paragraph{Visualization of Tokenized Sequence}
We compare a three second tokenized signal when $V = \{50, 100, 250\}$ to its original representation (far left) in Figure~\ref{Fig:token}.

\begin{figure}[!t]
\centering
\includegraphics[width=0.9\linewidth]{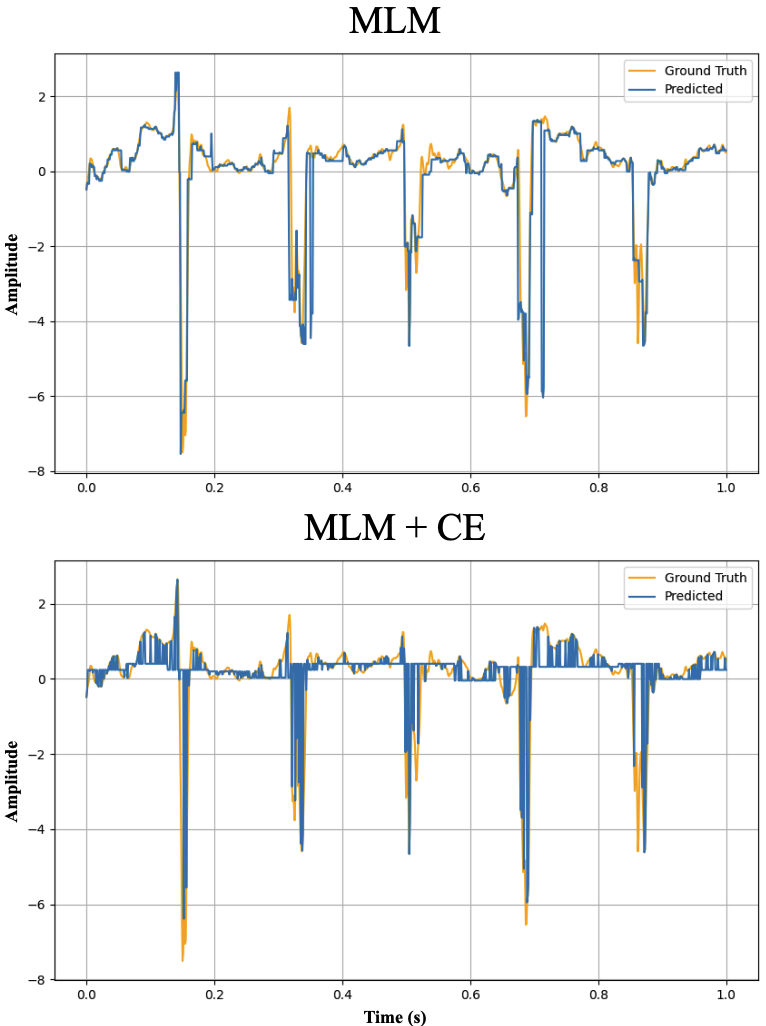}
\caption{Comparison of $L_{MLM}$ and $L_{MLM} + L_{CE}$ reconstructed AFib EGM signals with the BigBird model \citep{zaheer2021big} using our final learning objective.}
\label{Fig:interp}
\end{figure}

\paragraph{Visualization of Reconstructed EGM Signal}
We visualize the reconstructed EGM signal and compare between using only the MLM learning objective versus our full learning objective in Figure~\ref{Fig:interp}.
The reconstructed EGM signal using only the MLM learning objective follows the wave morphology more closely but results in lower AFib classification results. 
The reconstructed EGM signal using our full learning objective still closely resembles the signal's overall morphology, thus we accept this trade-off for the improvement in AFib classification.

\begin{figure*}[htp]
\centering
\includegraphics[width=1\linewidth]{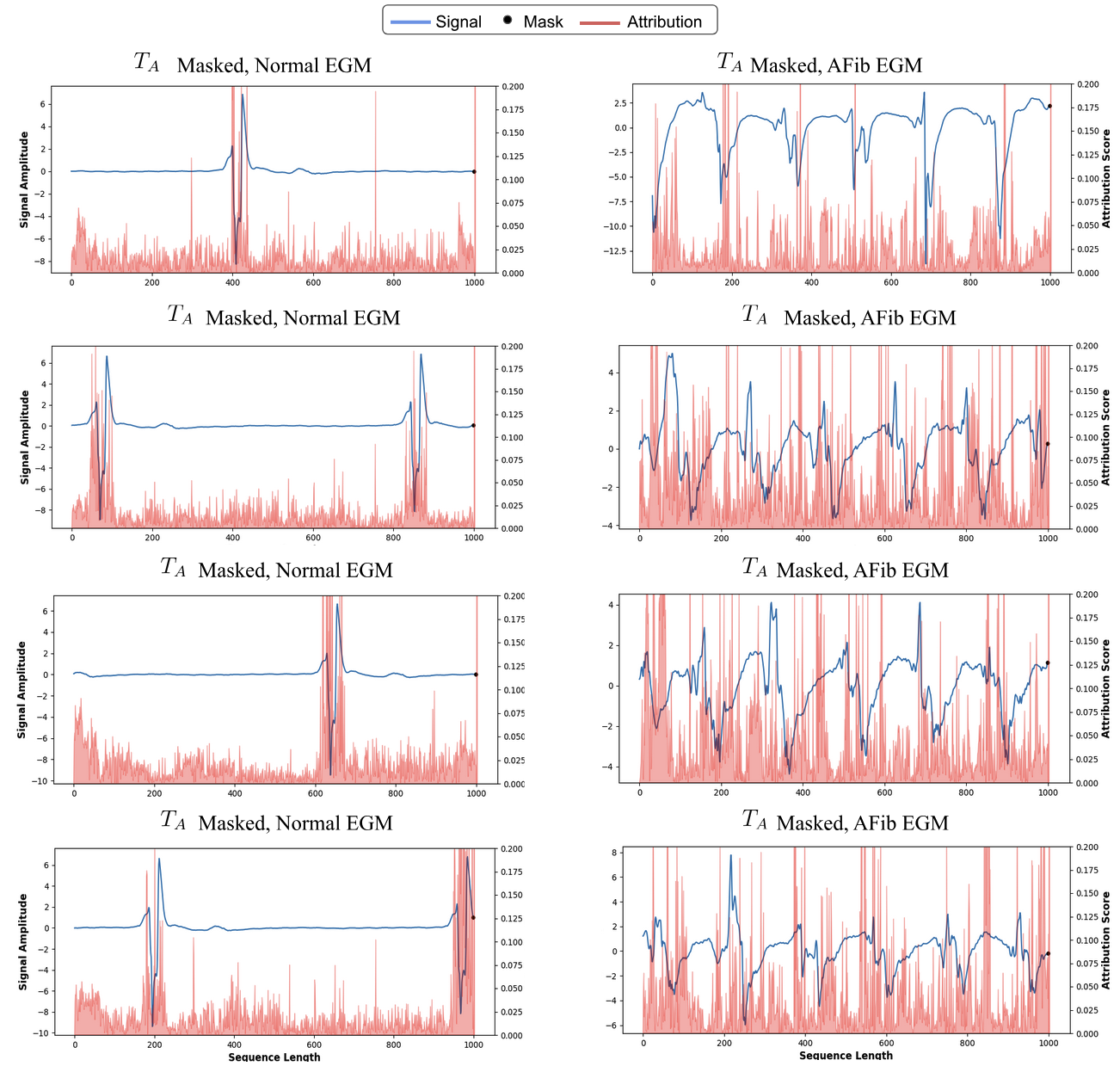}
\caption{We visualize the attribution scores of the finetuned checkpoint of our full learning objective using the BigBird model \citep{zaheer2021big}. We visualize randomly sampled normal and AFib EGMs with only masking the $T_A$ token for clinical significance.}
\label{Fig:attr2}
\end{figure*}

\paragraph{Visualization of Attribution Scores}
We randomly sampled 4 normal and 4 AFib EGMs and visualize the attribution score distributions in Figure~\ref{Fig:attr2}.
The visualizations are based off of the finetuned checkpoint of our full learning objective with the BigBird model \citep{zaheer2021big}. 
We also visualize the attribution score when the $T_A$ token is only masked out, due to the clinical significance.
From Figure~\ref{Fig:attr2}, we can observe that for normal EGMs,  the attribution scores are precisely localized around the fluctuating sections of the signal, which is aligned with the clinician's perspective.
For AFib EGMs, we can still observe a pattern formulating where the rise and fall of the attribution scores are aligned with the irregularly oscillating portions as well.

\vspace{-3mm}
\section{Details on Masked Language Models}
\paragraph{BigBird} The BigBird model employs a sparse attention mechanism to efficiently process long sequences. 
This approach is characterized by three types of attention patterns: random, windowed, and global. 
The random pattern enables long-range interactions modeled after the Erdős-Rényi graph. 
The windowed attention captures local context within a fixed-size window, while the global tokens are included to maintain overall sequence context. 
These elements are integrated into a modified attention calculation, reducing the complexity from quadratic to linear with respect to sequence length, thereby facilitating scalability for datasets with long textual information.
We use the pretrained weights `google/bigbird-roberta-base' provided by the authors \citep{zaheer2021big}. 

\paragraph{LongFormer} The Longformer model is also optimized for processing long sequences of data. 
\citet{beltagy2020longformer} introduce a sparse attention mechanism that combines sliding window and global attention, reducing complexity from quadratic to linear and enabling handling of up to 4096 tokens. 
The sliding window attention applies standard self-attention within a fixed-size window, ensuring local context is maintained. 
Global attention allows designated tokens to interact with the entire sequence. 
We use the pretrained weights `allenai/longformer-base-4096' provided by the authors \citep{beltagy2020longformer}.
\paragraph{Clinical BigBird and LongFormer} Clinical BigBird and Clinical LongFormer utilize the same architectures as BigBird and LongFormer, respectively.
The difference among these two models is that they are pretrained on 2 million clinical notes from the MIMIC-III dataset \citep{Johnson2016MIMICIIIAF}. 
We use the pretrained weights `yikuan8/Clinical-BigBird' and `yikuan8/Clinical-Longformer' for Clinical BigBird and Clinical LongFormer respectively, provided by the authors \citep{li2022clinicallongformer}.
\begin{table}[hbtp]
\floatconts
  {tab:sig_len}
  {\caption{Ablation Study on Signal Length $M$}}
  {\small
  \begin{tabular}{lcccccccc}
  \toprule
  & \multirow{2}{*}{\bfseries $M$}
  & \multicolumn{2}{c}{\bfseries Interpolation}
  & \multicolumn{1}{c}{\bfseries AFib Classification} 
  \\
  & &  \bfseries MSE $\downarrow$ 
  & \bfseries MAE $\downarrow$ 
  & \bfseries Accuracy \%
  \\
  \midrule
  & 4000 & 0.88 & 0.40 & 98.3\\
  & 3000 & 0.92 & 0.42 & \textbf{99.3}\\
  & 2000 & 1.18 & 1.04 & 97.4\\
  & 1000 & \textbf{0.80} & \textbf{0.37} & 99.2\\
  \bottomrule
  \end{tabular}}
\end{table}
\section{Additional Experiments}

\paragraph{Only MLM vs MLM + Classification head}
We compare our method with a traditional formulation of doing classification with a MLM. 
The traditional formulation is that given some input, in this case $[\text{CLS}] \oplus T_S \oplus [\text{SEP}]$, to the MLM, the output of it will be inputted into an additional linear layer for classification.
In our traditional formulation, we first finetune the MLM for the interpolation task then further finetune for AFib classification with an added linear layer. 
In \tableref{tab:mlm_lin}, we can see that our method is able to outperform the traditional formulation.
Therefore, we highlight the efficiency and superior performance of our method since we only require the MLM to be finetuned once.

\begin{table}[hbtp]
\floatconts
  {tab:mlm_lin}
  {\caption{Ablation study on MLM vs MLM + classification head}}
  {\resizebox{\columnwidth}{!}{
  \centering
  \begin{tabular}{lccccccc}
  \toprule
  \multirow{2}{*}{\bfseries Method}
  & \multicolumn{1}{c}{\bfseries AFib Classification} 
  \\
  & \bfseries Accuracy \%
  \\
  \midrule
  BigBird \citep{zaheer2021big} + Linear Layer& 98.7\\
  \textbf{Ours BigBird \citep{zaheer2021big}}& \textbf{99.2} \\
  \bottomrule
  \end{tabular}}}
\end{table}

\paragraph{Performance on Varying Signal Length $M$}
We compare the performance of our model when we vary the signal length $M$ in \tableref{tab:sig_len}, using the BigBird model \citep{zaheer2021big} with our final learning objective.
We observe that for interpolation, $M = 1000$ gives the best results, whereas for AFib classification, $M=3000$ does slightly better. 
In general, our model seems to be robust to varying signal length inputs.

\paragraph{Performance on Different Losses $L$}
We compare the performance of our model with three different loss $L$ formulations in Table~\ref{tab:dif_loss} : Our full learning objective $L_{MLM} + L_{AFib}$, only the MLM loss $L_{MLM}$, and only the classification loss $L_{AFib}$. Our reported results are utilizing the BigBird model \citep{zaheer2021big}.
We can see a clear performance tradeoff between the different learning objectives.
Notably, when we add the $L_{AFib}$ with $L_{MLM}$ we can see that the accuracy rises while the MSE and MAE scores increases.
Additionally, we visualize in Figure~\ref{Fig:interp} and observe that the reconstructed EGM signal using our full learning objective is comparable to only using $L_{MLM}$ while increasing the accuracy, thus we deem this tradeoff acceptable.

\begin{table}[hbtp]
\floatconts
  {tab:dif_loss}
  {\caption{Ablation study on different losses $L$.}}
  {\scriptsize
  \setlength{\tabcolsep}{3pt}
  \begin{tabular}{lcccc}
  \toprule
  & \multirow{2}{*}{\bfseries $L$}
  & \multicolumn{2}{c}{\bfseries Interpolation}
  & \multicolumn{1}{c}{\bfseries AFib Classification} 
  \\
  & &  \bfseries MSE $\downarrow$ 
  & \bfseries MAE $\downarrow$ 
  & \bfseries Accuracy \%
  \\
  \midrule
  & $L_{AFib}$ & 36.79 & 4.17 & 63.0\\
  & $L_{MLM}$ &\textbf{0.44} & \textbf{0.16} & 96.7\\
  & $L_{MLM} + L_{AFib}$ & 0.80 & 0.37 &\textbf{99.2} \\
  \bottomrule
  \end{tabular}}
\end{table}

\paragraph{Performance on Different Bin Levels $V$.}
We compare the performance of our model with different bin levels $V$ in Table~\ref{tab:dif_bin}. We can observe that although the MSE and MAE scores are very good for lower $V$, the accuracy takes a toll. 
Through this empirical analysis, we found that $V=250$ to be the most appropriate number to maintain good performance across interpolation and classification.
Our reported results are utilizing the BigBird model \citep{zaheer2021big}.

\begin{table}[hbtp]
\floatconts
  {tab:dif_bin}
  {\caption{Ablation study on different bin levels $V$.}}
  {\small
  \setlength{\tabcolsep}{3pt}
  \begin{tabular}{lcccc}
  \toprule
  & \multirow{2}{*}{\bfseries $V$}
  & \multicolumn{2}{c}{\bfseries Interpolation}
  & \multicolumn{1}{c}{\bfseries AFib Classification} 
  \\
  & &  \bfseries MSE $\downarrow$ 
  & \bfseries MAE $\downarrow$ 
  & \bfseries Accuracy \%
  \\
  \midrule
  & 50 & \textbf{0.40} & 0.24 & 64.4\\
  & 100 &0.42 & 0.19 & 64.2\\
  & 150 & 0.42 & \textbf{0.17} &68.43 \\
  & 200 & 0.45 & \textbf{0.17} &75.41 \\
  & 250 & 0.80 & 0.37 &\textbf{99.2} \\
  \bottomrule
  \end{tabular}}
\end{table}

\paragraph{Performance on Fitting Previous Methods to Our Data}
For a fairer comparison to other methods, we implemented previous methods proposed in \citet{alhusseini_2020_machine} and \citet{tang_2022_machine} to our data and present the results in Table~\ref{tab:prev}. 
Although, we do want to note that for \citet{alhusseini_2020_machine}, they utilize a basket catheter with 64 electrodes to collect their data.
Additionally, they create a spatial heatmap that represents a 8 x 8 grid of the 64 electrodes as input to the CNN. 
In our setting, we only use one out of the 48 electrodes of our Octoray catheter as input to the CNN for a fair comparison.
We can see from the table that our method is able to perform better compared with previous methods.

\begin{table}[hbtp]
\floatconts
  {tab:prev}
  {\caption{Results on classification task utilizing models proposed in \citet{alhusseini_2020_machine} and \citet{tang_2022_machine}.}}
  {\resizebox{\columnwidth}{!}{
  \centering
  \begin{tabular}{lccccccc}
  \toprule
  \multirow{2}{*}{\bfseries Method}
  & \multicolumn{1}{c}{\bfseries AFib Classification} 
  \\
  & \bfseries Accuracy \%
  \\
  \midrule
  CatBoost \citep{tang_2022_machine}& 90.6\\
  CNN \citep{alhusseini_2020_machine} & 91.6\\
  K-Means \citep{alhusseini_2020_machine} & 66.5 \\
  KNN \citep{alhusseini_2020_machine}& 82.5\\
  LDA \citep{alhusseini_2020_machine} &65.5\\
  SVM \citep{alhusseini_2020_machine}& 64.6\\
  \textbf{Ours BigBird \citep{zaheer2021big}}& \textbf{99.2 }\\
  \bottomrule
  \end{tabular}}}
\end{table}

\end{document}